\documentclass[11pt]{article}

\usepackage[final]{latex/acl}

\usepackage{times}
\usepackage[T1]{fontenc}
\usepackage{array}
\usepackage{enumitem}
\usepackage{todonotes}

\usepackage{pifont}
\definecolor{mediumgreen}{HTML}{13B015}
\newcommand{\tick}{{\color{mediumgreen} \ding{51}}}
\newcommand{\cross}{{\color{red} -}}
\definecolor{g}{RGB}{0,0,204}     % Dark blue for guru
\definecolor{l}{RGB}{0,102,0}    % Dark green for laghu
\definecolor{h}{RGB}{153,0,153} % Purple for hrasva
\definecolor{de}{RGB}{204,0,0}  % Red for deergha
\newcommand{\ourdataset}{{\sl Chandomitra}~}

\usepackage{multicol}
\usepackage{makecell}

\colorlet{an1}{blue!50}
\colorlet{an2}{orange}
\colorlet{an3}{yellow!80!orange}
\colorlet{anbox1}{yellow!50!orange}
\colorlet{anbox2}{cyan}

\newcommand{\emb}[1]{\texttt{emb}(#1)}

\usepackage{microtype}

\usepackage{inconsolata}

\usepackage{graphicx}
\usepackage{enumitem}
\usepackage{booktabs}
\usepackage{longtable}
\usepackage{amssymb}
\usepackage{amsmath}
\usepackage{array}
\usepackage[linesnumbered,ruled,vlined]{algorithm2e}

\usepackage{tcolorbox} % For custom boxed layout
\usepackage{caption} % Needed for captionof
\newtcolorbox{AlgorithmBox}[1][]{
  enhanced,
  sharp corners,
  colback=white,
  colframe=black,
  fonttitle=\bfseries,
  title=Algorithm: Constrained Decoding for Sanskrit Poetry Generation,
  #1
}

\usepackage{adjustbox}
\usepackage{float}
\usepackage{authblk}
\usepackage{textcomp}
\usepackage{tipa}
\usepackage{skt}

\title{Chandomitra: Towards Generating Structured Sanskrit Poetry from Natural Language Inputs}

\author{Manoj Balaji Jagadeeshan\thanks{Equal contribution as first authors.}$^{1}$ Samarth Bhatia$^{*2}$ Pretam Ray\thanks{Equal contribution as second authors.}$^{1}$ \\ Harshul Raj Surana$^{\dagger1}$ Akhil Rajeev P$^{3}$ Priya Mishra$^{4}$ \\Annarao Kulkarni$^{3}$ Ganesh Ramakrishnan$^{4}$ Prathosh AP$^{2}$ Pawan Goyal}
\affil[1]{Indian Institute of Technology, Kharagpur}
\affil[2]{Indian Institute of Science, Bangalore}
\affil[3]{Indian Heritage Language Computing Group, Centre for Development of Advanced Computing (C-DAC), Bangalore}
\affil[4]{Indian Institute of Technology, Bombay}

\begin{document}
    \maketitle
\begin{abstract} 
% \textbf{Placeholder Abstract}
% Recent advances in large language models (LLMs) have significantly improved natural language generation, including creative tasks like poetry composition. However, most progress remains concentrated in high-resource languages. This raises an important question: \textit{Can LLMs be adapted for structured poetic generation in a low-resource, morphologically rich language such as Sanskrit?} 
% In this work, we introduce a dataset designed for translating English prose into structured Sanskrit verse, with strict adherence to classical metrical patterns, particularly the \texttt{Anu\d{s}\d{t}ubh} meter. We evaluate a range of generative models—both open-source and proprietary—under multiple settings. Specifically, we explore constrained decoding strategies and instruction-based fine-tuning tailored to metrical and semantic fidelity.
% Our decoding approach achieves over 99\% accuracy in producing syntactically valid poetic forms, substantially outperforming general-purpose models in meter conformity. Meanwhile, instruction-tuned variants show improved alignment with source meaning and poetic style, as supported by human assessments, albeit with marginal trade-offs in metrical precision.
Text Generation has achieved remarkable performance using large language models. It has also been recently well-studied that these large language models are capable of creative generation tasks but prominently for high-resource languages.
This prompts a fundamental question: \textit{Is there a way to utilize these (large) language models for structured poetry generation in a low-resource language, such as Sanskrit?} We present \ourdataset, an English input to structured Sanskrit Poetry translation dataset, specifically adhering to the {\sl Anushtubh} meter. We benchmark various open and closed models, and scrutinize specialized techniques such as constrained decoding and instruction fine-tuning, for the proposed task. 
Our constrained decoding methodology achieves 99.86\% syntactic accuracy in generating metrically valid Sanskrit poetry, outperforming GPT-4o (1-shot: 31.24\%). Our best-performing instruction-tuned model, on the other hand, performs better in semantic coherence with the English input, at the expense of slightly lower syntactic accuracy. Human evaluation further reveals that instruction fine-tuned model is better able to capture the poetic aspects. Data\footnote{\href{https://huggingface.co/collections/sanganaka/chandomitra}{https://huggingface.co/collections/sanganaka/chandomitra}} and Code\footnote{\href{https://github.com/sanganaka-iitkgp/chandomitra}{https://github.com/sanganaka-iitkgp/chandomitra}} are available.

\end{abstract}

\section{Introduction}

Language constitutes a key aspect of human intelligence; among its various forms, poetry stands out as a beautiful, creative genre that conveys emotions and ideas with relative brevity. Poetry is universally captivating across diverse nations, ethnicities, and cultures, significantly influencing the evolution of human civilization.

Large language models (LLMs) have made substantial progress in various domains including complex task-solving \cite{feng-etal-2024-large}, reasoning \cite{wei2023chainofthoughtpromptingelicitsreasoning}. Studies also have shown that large language models such as \texttt{GPT-3} \cite{brown2020languagemodelsfewshotlearners}, \texttt{LLaMa} \cite{touvron2023llamaopenefficientfoundation},  etc., have advanced the state of the art in several natural language generation tasks such as text summarization \cite{zheng-etal-2020-controllable} and machine translation \cite{nllbteam2022languageleftbehindscaling}. The use of these models for creative generation tasks, such as creative story or poetry generation, is, however, limited to a small number of works \cite{chakrabarty-etal-2022-help, yang2020generatingthematicchinesepoetry}.

Most prior research has focused on high-resource languages, including English or Chinese, where the source and target are in the same language. Sanskrit, a classical language \citep{coulson1976sanskrit}, primarily comprises sentences structured as verses in its pre-classic and classic literature. Most existing Sanskrit corpora are composed in verse form \cite{dcs}. While there have been works on identification of meter from a Sanskrit poetry (verse) \cite{terdalkar2022chandojnanamsanskritmeteridentification}, there are only couple of works related to Sanskrit poetry \cite{krishna-etal-2019-poetry, das2025therellmsoutperformsmaller}, that too, given the corresponding Sanskrit prose format, thus tackling it as a reordering task given the meter constraints. 

This study presents a dataset \ourdataset, derived from the Valmiki Ramayana corpus\footnote{\href{https://www.valmiki.iitk.ac.in}{https://www.valmiki.iitk.ac.in}}. %\tick\pg{Link here}
In our setting, the source (English) and target (Sanskrit) are in different languages, and we focus on generating Sanskrit verses in \texttt{Anu\d{s}\d{t}ubh} Chandas form, which is a particular metrical form of Sanskrit poetry. Thus, this task has a unique setting that requires both translation as well as structured poetry generation. \\
Despite being provided with metrical constraints and examples in the $k$-shot prompt (with best performance for $k=1$ when $k\in [0,3]$),  \texttt{GPT-4o} achieves only 31.24\% syntactic correctness (52.84\% being partially correct) on in-domain samples, and 21.84\% on out-of-domain samples. This highlights the challenges of structured poetry generation in Sanskrit, a task that requires strict adherence to rhythm and meter, and motivates the need for more specialized approaches. \\
We propose multiple possible approaches to address the challenges of structured Sanskrit poetry generation. First, to enforce metrical constraints during inference, we design a constrained decoding strategy utilizing \texttt{NLLB-dist-1.3B} that counts syllables and employs precompiled regex according to the Sanskrit \texttt{Anu\d{s}\d{t}ubh} meter to generate metrically valid outputs with an accuracy of 99.86\%. %\tick\pg{Revise the score.} 
However, this comes at a cost of reduced semantic similarity to the original input sentence. \\ 
Next, we enhance the models' adherence to metrical restrictions via instruction fine-tuning \cite{wei2022finetunedlanguagemodelszeroshot}, providing the detailed instructions and constraints in the prompt. We find that the Instruction-finetuned models balance syntax and semantics: \texttt{Phi-4-14B} and \texttt{Mistral-Nemo-2407-12B} achieve syntactic accuracies of 57.42\% and 52.5\%, and semantic scores of 67.29 and 68.05. \\
Our key contributions are as follows:\\
    (a) We introduce \ourdataset, a novel translation-cum-poetry generation dataset for structured Sanskrit \texttt{Anu\d{s}\d{t}ubh} poetry from English inputs. \\
    (b) We propose syntactic and semantic evaluation metrics for the task. For semantic evaluation in this cross-lingual prose-to-poetry setting, we collect human judgments for \textasciitilde1k English-Sanskrit pairs. We experiment with candidate metrics and select the one with the best correlation, enabling a nuanced assessment of meaning preservation beyond syntax.\\
    (c) We benchmark models including \texttt{GPT-4o}, and enhance performance via constrained decoding and instruction fine-tuning. We further perform human evaluation for the best performing \texttt{NLLB-distilled-1.3B} and \texttt{Mistral-Nemo-2407-12B} models. Human evaluation shows that while \texttt{NLLB-dist-1.3B} (constrained decoding) ensured metrical correctness, it led to grammatical discontinuities, lowering ratings. \texttt{Mistral-Nemo-2407-12B}, on the other hand, produced semantically and poetically superior verses.

\section{Relevant Background}
\label{p:rules}
\subsection{Sanskrit Poetry}

Sanskrit poetry, a cornerstone of Indian literary tradition, is deeply rooted in metrical compositions known as Chandas. 
The study of Chandas ensures precision in poetic composition and facilitates memorization and recitation.

The term Chandas refers to the metrical framework of Sanskrit poetry. It regulates the arrangement of syllables into specific patterns of \textcolor{l}{light (laghu)} and \textcolor{g}{heavy (guru)} syllables within a line or stanza. These patterns create rhythm and enhance the aesthetic appeal of the verse. Each metrical unit is called a \texttt{p\={a}da}, and a verse typically comprises four \texttt{p\={a}das}. The science of Chandas is elaborated in ancient texts like \texttt{Pingala's Chandaḥ\'{s}\={a}stra} \cite{pingala_chhanda}.

\noindent\textbf{Constituents of Chandas.}

\noindent\texttt{P\={a}da}: A quarter of a stanza, each \texttt{p\={a}da} contains a fixed number of syllables based on the meter.

\noindent Laghu (Light Syllable): A short syllable, generally with a short vowel and no consonant cluster following it.

\noindent Guru (Heavy Syllable): A long syllable, characterized by a long vowel, a consonant cluster, or ending with an \texttt{anusv\={a}ra/visarga}.

%\end{enumerate}
 
More details can be found in \autoref{app:ext-chandas}.

\subsection{Anu\d{s}\d{t}ubh Chandas}
\label{anushtubh}
The \texttt{Anu\d{s}\d{t}ubh} meter is one of the most prevalent meters in Sanskrit literature, used extensively in texts like the Mahabharata, Ramayana, and Bhagavad Gita. It consists of 32 syllables divided into four \texttt{p\={a}das}, each containing eight syllables\footnote{The definition provided in sanskrit is available in Appendix \autoref{anushtubh-def}
\label{sec:anushtubh-rules}}.

\begin{figure}[ht]
    \centering
    \input{rules-anushtup-color-arxiv}
    \caption{An example of \texttt{Anu\d{s}\d{t}ubh} Chanda. Each pada has 8 syllables. \colorbox{anbox1}{$\cdot$}: odd pada; \colorbox{anbox2}{$\cdot$}: even pada.}
    \label{fig:rules-anushtup}
\end{figure}

The \texttt{Anu\d{s}\d{t}ubh} meter follows these precise rules:

\begin{itemize}[leftmargin=*, itemsep=0pt]
    \item Each \texttt{p\={a}da} consists of exactly eight syllables
    \item The sixth syllable in every \texttt{p\={a}da} must be \textcolor{g}{guru} (heavy)
    \item The fifth syllable is always \textcolor{l}{laghu} (light)
    \item For the seventh syllable:
    \begin{itemize}[leftmargin=*, itemsep=0pt]
        \item Must be \textcolor{de}{deergha} (long) or \textcolor{g}{guru} (heavy) in odd-numbered \texttt{p\={a}das} (1st and 3rd)
        \item Must be \textcolor{h}{hrasva} (short) in even-numbered \texttt{p\={a}das} (2nd and 4th)
    \end{itemize}
\end{itemize}

\subsection{Poetry Generation in Other Languages}
Early approaches to poetry generation used rule-based, statistical, and RNN-based models \citep{poetryme,jiang-zhou-2008-generating,zhang-lapata-2014-chinese}. Transformer models like \texttt{GPT} show mixed results-while fine-tuned variants of \texttt{GPT-2} can produce stylistically rich poetry \cite{bena2020introducingaspectscreativityautomatic}, zero-shot outputs often lack meter and rhyme \cite{köbis2020artificialintelligenceversusmaya,wockener-etal-2021-end}. To address this, \texttt{BYGPT5} introduces a token-free architecture tailored to poetic forms, outperforming larger models \citep{belouadi-eger-2023-bygpt5}. Fine-tuned \texttt{GPT-3} can emulate author styles \citep{Sawicki2023}, whereas \texttt{GPT-3.5} and \texttt{GPT-4} underperform without fine-tuning. Post-editing methods have also been proposed to improve stylistic fidelity \citep{ma-etal-2023-yu}
While prior work has explored prose-to-poetry transformations in Sanskrit \citep{krishna-etal-2019-poetry}, we present the first dataset and model targeting the inverse task: generating structured Sanskrit verse from English prose.

\subsection{Problem Definition}
Formally, given an input sentence (prose) in English $x$, we need to generate it's corresponding output (poetry) $\hat{y}$ in Sanskrit s.t. $\hat{y}$ follows the syntactic rules of the \texttt{Anu\d{s}\d{t}ubh} chanda as detailed in \autoref{anushtubh} and preserves the meaning of the input $x$. The ground-truth Sanskrit poetry, if required, is denoted by $y$.

In other words, this task implies \textit{translation and structured poetry generation at the same time}. 

\section{The \ourdataset Dataset}
To facilitate this research, we construct an initial dataset derived from the Valmiki Ramayana corpus\footnote{\url{https://www.valmiki.iitk.ac.in/}}, retaining only verses in the \texttt{Anu\d{s}\d{t}ubh} meter using \texttt{Chandoj\~{n}\={a}nam}\footnote{\url{https://github.com/hrishikeshrt/chanda/}}, a tool that lets us find the meter of Sanskrit poetry. 
This curated dataset forms the basis for training and evaluating our models. Table \ref{tab:dataset-stat} gives the statistics of the \ourdataset dataset, consisting of 9,727 verses in \texttt{Anu\d{s}\d{t}ubh} meter, along with the English translation. 
One such pair is shown here, with more examples given in \autoref{sec:dataset-examples}.

\noindent\fbox{
\begin{minipage}{\dimexpr\linewidth-2\fboxsep-2\fboxrule\relax}
    
\textbf{English Translation (Prose)}:\\
Effulgent Rama looked at Khara who stood with a mace in hand minus his chariot and said to him first in a gentle voice and then harshly 

\textbf{Ground Truth Sanskrit (\texttt{Anu\d{s}\d{t}ubh} Poetry)}: \\
{{\skt Ka:=\ZH{-6}{M} tua ;Y2a;va:=+TMa .=+a;ma;ea
ga;d;a;pa;a;Y5a;Na;ma;va;\ZH{0}{i0//////Y7}a;s1Ta;ta;m,a \ZS{12}@A \\
m\ZV{2}{x}a;d\ZH{-10}{u};pUa;vR2a ma;h;a;tea:ja;aH\ZS{4} :pa:r8+:SMa
va;a;ky2+.a;ma;b.ra;vi6a;a;t,a \ZS{12}@A\ZS{6}@A}}
\end{minipage}
}

\begin{table}[h]
\small
    \centering
    \resizebox{\linewidth}{!}{
    \begin{tabular}{lcc}
        \toprule
        \textbf{Dataset} & \textbf{\# Train samples}& \textbf{\# Test samples}\\
        \midrule
        \ourdataset & 8306 & 1421  \\ 
        OOD & $-$ & 520 \\ 
        \bottomrule
    \end{tabular}
    }
    \caption{Dataset statistics for \ourdataset and out-of-domain (OOD) test set.}
    \label{tab:dataset-stat}
    
\end{table}

In addition, to assess the generalizability of our models, we prepare an out-of-domain dataset. Specifically, we included Raghuvansh by Maharshi Kalidas (468 verses) and Leelavati by Acharya Bhaskaracharya (52 verses). Although these texts belong to different poetic traditions compared to the Valmiki Ramayana corpus, the selected verses are also in the \texttt{Anu\d{s}\d{t}ubh} meter, ensuring consistency in the metrical structure across datasets. 
 
The total number of out-of-domain samples amounts to 520 pairs.

\section{Proposed Solution}

To generate Sanskrit poetry conforming to the strict \texttt{Anu\d{s}\d{t}ubh} metrical constraints, we adopt two complementary strategies aligned with the strengths of distinct model families. For raw translation models, inherently sequence-to-sequence and lacking instruction-following abilities, we use constrained decoding to enforce syllabic and structural rules during generation. This guides the decoder toward metrically valid outputs without altering the architecture. For instruction-following models, we apply instruction-fine-tuning (IFT) to internalize syntactic and semantic patterns of Sanskrit verse via curated prompts and data. These models interpret explicit constraints in natural language, making them apt for controllable generation when fine-tuned on metrically aligned samples. These dual approaches exploit architectural strengths $-$ constrained decoding for symbolic control and IFT for contextual fluency $-$ yielding metrically faithful and semantically coherent verse.

Next, we describe the proposed approaches for structured Sanskrit \texttt{Anu\d{s}\d{t}ubh} generation below.

\subsection{Constrained Decoding}

To ensure that generated Sanskrit poetry adheres to the strict syllabic patterns of metrical form of \texttt{Anu\d{s}\d{t}ubh}, we employ a constrained decoding strategy tailored for symbolic control during inference. This approach, detailed in \autoref{alg:constrained_decoding}, enables metrically faithful generation without requiring any modification to the underlying language model.

At each generation step, the LM provides probabilities for each token in its vocabulary. To maintain computational efficiency and semantic relevance, we select the top $k$ tokens as a possible next token, and verify which of the resulting syllable weight sequences satisfy our constraints. Precompiled regex filters are used on the syllable weights to enforce the patterns allowed for each pada.

For instance, the filter for the first pada of an \texttt{Anu\d{s}\d{t}ubh} verse (8 syllables with the 5th and 6th syllables being laghu and guru, respectively, and the 7th syllable being guru) would be \verb|"^.{4}lgg.$"| (signifying that the first 4 syllable weights are \verb|.| (unconstrained) and the next three can only be \verb|l|(laghu), \verb|g|(guru) and \verb|g|(guru), in that order, followed by the 8th one being unconstrained again. The \verb|^| and \verb|$| signify the beginning and end of a line. Sequences that do not match the filter are discarded. We then use greedy sampling (we also test with other sampling techniques) to select one from the tokens that satisfy the constraint, pass it to the LLM for next token prediction, repeating until a length constraint is met (such as the 32 syllable rule in \texttt{Anu\d{s}\d{t}ubh}).

While the algorithm has a similar form to general constrained decoding algorithms, the implementation details here demand effort to get precisely right. Particularly, we had to get strict, fine-grained control over the generation and control exactly which tokens can be predicted near pada boundaries, refined syllable-pada division checks, preventing certain duplicate tokens, when to allow modifier-only tokens (such as \texttt{matras}), etc.

\begin{algorithm}
% I know this gives an error but let it be, it ignores it and runs with it.
\caption{Constrained Decoding for Sanskrit Poetry Generation
%\pg{In the algo, the sampling part is not mentioned.} {\color{cyan} - Fixed}
}\label{alg:constrained_decoding}
\resizebox{0.877\linewidth}{!}{
\begin{minipage}{1.5\linewidth}
\SetKwInOut{Input}{Input}
\SetKwInOut{Output}{Output}

\Input{LLM, metrical constraints as regular expression filters}
\Output{Generated Sanskrit poetry adhering to metrical constraints}

$|V| \gets$ Vocab size of the LLM \quad
$k \gets 25$

\While{not end of generation}{
    
    \textbf{Step 1:} Generate token probabilities from the LLM\quad
    $P \in \mathbb{R}^{|V|} \gets \textbf{LLM}(\text{input})$\;
    \textbf{Step 2:} Sample top-$k$ possible tokens\quad
    Top-$k$ Indices: $I_k \gets \texttt{topk}(P, k)$\;
    \For{each token at index $I_i$ in top-$k$ token indices $I_k$}{
        Token $T_i \gets \text{vocab}[I_i], \quad$
        $O_{\text{temp}} \gets \{\text{...input tokens}\}$\;
        \textbf{Step 3:} Temporarily add $T_i$ to output $O_{\text{temp}}$
        $\quad O_{\text{temp}} \gets O_{\text{temp}} || T_i$\;
        \textbf{Step 4:} Calculate the new syllable weighting for $O_{\text{temp}}$
        $\quad  w_{\text{temp}} \gets \text{syllable-weights}(O_{\text{temp}})$\;
        \textbf{Step 5:} Match $w_{\text{temp}}$ against pre-compiled regex filters of the same length\;
        \If{$w_{\text{temp}}$ does not match the filters}{
            \textbf{Step 6:} $P[I_i] \gets -\inf$
        }
        \textbf{Step 7:} Check length rules (e.g., end of pada or verse)\;
        % \If{token satisfies length rules}{
        %     \textbf{Step 8:} Break the loop and return $T_i$ as the chosen token.\;
        % }
        % \Else{
        % }
    }
    \If{no token matches the constraints}{
        \textbf{Step 8:} Increase $k$ and repeat the process from Step 1\;
    }
    \textbf{Step 9:} Return modified probabilities for further sampling (greedy/nucleus/etc.)
}
\end{minipage}
}
\end{algorithm}
% }

\subsection{Instruction Fine-tuning}

Previous works have demonstrated the effectiveness of improving a model's in-context learning abilities through prompts \citep{zhong-etal-2021-adapting-language, weir2022oneshotlearningdemonstrationhierarchical} or by leveraging demonstration examples \citep{min2022filmfollowinginstructionslanguage}. Building on this, we perform instruction fine-tuning (IFT) on a set of quantized decoder-only language models using our curated \ourdataset{} training corpus. This includes models from high-performing families such as \texttt{LLaMA-3}, \texttt{Mistral-Nemo}, and \texttt{Phi-4} \citep{sreenivas2024llmpruningdistillationpractice}. 
 
Through carefully constructed prompts and fine-tuning data, anchored in metrical and lexical rules, these models learn to respect poetic constraints specified in natural language, making them well-suited for controllable generation. 
\section {Experiments}

\subsection{Evaluation metrics}
\label{sec:metrics}

We cannot use standard metrics such as BLEU \cite{bleu}, chrF \cite{popovic-2015-chrf}, or ROUGE \cite{lin-2004-rouge} because these metrics are based on the ordering of the words in the output being the same as the ordering in the ground truth. However, these metrics do not evaluate the task we are trying to solve.

\autoref{tab:standard-metric} shows the results for the above standard metrics on randomly sampled 30 sentences from \ourdataset test split. Clearly, the normal translation-based metrics do not do justice to this task.

\begin{table}[!thb]
    \centering
    \resizebox{\linewidth}{!}{
    \begin{tabular}{l|ccc}
        \hline
        \textbf{Type} & \textbf{Approach}& \textbf{bleu} & \textbf{chrF} \\
        \midrule
        \texttt{Llama3.1-8B-Instruct} & Instruction Fine-Tuning  & 0.0 & 17.75 \\ 
        \texttt{NLLB-dist-1.3B-FT} & Constrained Decoding  & 0.0 & 17.09 \\ 
        \texttt{GPT-4o} & 1-shot prompt & 0.0 & 19.14 \\ \hline
    \end{tabular}
    }
    \caption{Issue with the standard metrics -- \ourdataset test dataset}
    \label{tab:standard-metric}
\end{table}

Thus, we choose to analyze generated poetry through:

\noindent \textbf{Syntactic metrics}: whether the generation follows metrical constraints properly\\
\noindent \textbf{Semantic metrics}: whether the generation conveys the same meaning as the English input

\subsubsection{Syntactic Metrics}

\noindent \textbf{Full \texttt{Anu\d{s}\d{t}ubh}}: The percentage of generated samples which completely adheres to the rules of \texttt{Anu\d{s}\d{t}ubh}. We use \verb|skrutable| for this \cite{skrutable}, which allows us to syllabify text and identify it's meter in a structured manner.

\noindent \textbf{Partial \texttt{Anu\d{s}\d{t}ubh}}: The percentage of samples that contain the required 32 syllables in length. This is a more relaxed metric and includes samples that follow Anushutbh rules, so this metric will always be $\geq$ (in magnitude) the Full \texttt{Anu\d{s}\d{t}ubh} metric.

\subsubsection{Semantic metrics}
\label{sec:semantic-similarity-methods}

It is important for the generated poetry to convey the same meaning as the input text. Thus, we need an automated metric to quantify semantic similarity between English (prose) and Sanskrit (poetry).
To this end, we compared three methods and chose the best one based on their correlation to human evaluation of semantic similarity. 

\textbf{Human Annotation}. We constructed a dataset of 1000 samples taking English input and Sanskrit poetry pairs from the \ourdataset dataset as well as generated poetry of various models. Then, we distributed 500 each to two human annotators and asked them to give a score between 1-5 to each (generated) poetry following the scale given in \autoref{app:human-annotation-metriceval} asking if the original English input intent is preserved.

\textbf{Semantic metrics comparison.}
Let $x_i$ be the $i$-th input sample from the dataset $D$, and $\hat{y}_i$ be it's corresponding output $\forall i \in \{1,2,\cdots|D|\}$.

We consider the following possible candidates for semantic similarity computation. 

\newcommand{\metricA}{{\color{magenta}\textbf{A}}}
\newcommand{\metricB}{{\color{cyan}\textbf{B}}}
\newcommand{\metricC}{{\color{green}\textbf{C}}}
\newcommand{\metricD}{{\color{orange}\textbf{D}}}
\textbf{Cross-lingual metrics}
\begin{itemize}[leftmargin=*]
    \item[\metricA] Fine-tuning sentence transformer models and calculating the \emph{linearized} cosine similarity between embeddings $\emb{\cdot}$ of $x_i$ and $\hat{y}_i$. More details pertaining to the fine-tuning of these models are given later.
    $$\text{score}_i = \frac{\pi - \cos^{-1}(\cos(\emb{x_i}, \emb{\hat{y}_i}))}{\pi}$$
    We linearize the cosine similarity to improve metric interpretability and comparability, so that it is easier to compare directly between different models/approaches.
    
    \item[\metricB] Using the margin-based scoring method introduced in \citet{artetxe-schwenk-2019-margin}. We calculate the embeddings $\emb{\cdot}$ using a finetuned sentence-transformer model trained in \metricA, and use "ratio" as the margin function. 
    $$\text{score}_i = \frac{ \cos(\emb{x_i}, \emb{\hat{y}_i})}{\frac{1}{2}(\text{Avg}(\emb{x}) + \text{Avg}(\emb{\hat{y}}))}$$ \\ $$\text{where}, \text{Avg}(t) = \frac{1}{k}\sum_{z \in \text{NN}_k(t)}\cos(t,z)$$
    $\text{Avg}(t)$ is the average cosine similarity of $t$ with its $k$ nearest neighbors ($\text{NN}_k$) in the \textbf{other} language.
\end{itemize}

\textbf{Monolingual metriccs}
\begin{itemize}[leftmargin=*]
    \item[\metricC] Translating the generated poetry back to English using NLLB-3.3B and calculating the BERTScore between the English input and the English translation of the generated Sanskrit \citet{zhang2020bertscoreevaluatingtextgeneration}
    \item[\metricD] Using fine-tuned sentence transformer models trained in \metricA\ to calculate the cosine similarity between embeddings $\emb{y_i}$ (the Sanskrit Ground Truth verse) and $\emb{\hat{y}_i}$.
\end{itemize}

\textbf{Fine-Tuning Sentence Transformer Models for Embedding}
\label{p:sentence-transformer-similarity}

Given the absence of models specifically trained on Sanskrit data, we fine-tune existing models using the \texttt{mitrasa\d{m}graha} dataset\cite{nehrdich2026mitrasamgrahacomprehensiveclassicalsanskrit}, followed by assessing the models' translation accuracy on a holdout test set. The fine-tuning makes use of the multiple negatives ranking loss. The translation accuracy is calculated by \verb|TranslationEvaluator| from the \verb|sentence-transformers| package \cite{reimers-2019-sentence-bert}, which internally checks pairwise if a sample in English has the highest similarity to that same sample in Sanskrit or another Sanskrit sample. This is done in both directions and the average translation accuracy results are presented in \autoref{tab:semantic_sim}. Based on these results, we choose \texttt{BAAI/bge-m3} as the best embedding model for metric choices \metricA, \metricB\ and \metricD.

\begin{table}[h]
    \centering
    \resizebox{\linewidth}{!}{
    \begin{tabular}{lcr}
        \toprule
        Model & \# of Params (M) & Mean Acc.$\dagger$ \\
        \midrule       
        \texttt{paraphrase-multilingual-MiniLM-L12-v2} & 118&85.39\\
        \texttt{LaBSE} & 470&88.61\\
        \texttt{Alibaba-NLP/gte-multilingual-base} & 305&90.52\\
        \texttt{BAAI/bge-m3} & 560&\textbf{92.88}\\
        \texttt{FacebookAI/xlm-roberta-large} & 560&92.08\\
        % Facebook/MEXMA& 560& {\color{red} TODO} \\
        \bottomrule
    \end{tabular}
    }
    \caption{Evaluation of sentence-transformers models trained on a Sanskrit dataset.\\ $\dagger$ Mean Acc. refers to the mean translation accuracy (mean of English to Sanskrit, and Sanskrit to English). }
    \label{tab:semantic_sim}
\end{table}

\textbf{Correlation with Human Annotation.} We see in \autoref{tab:human_corr} that linearized cosine similarity based semantic similarity metric (\metricA) has a high positive correlation with human evaluation. 

For \metricA, the linearization does not have an effect on the Spearman correlation as it is based on the relative ranking, which does not change with this operation. However, the Pearson correlation does increase slightly (by 0.02). For further analysis, we had also evaluated other sentence transformer models against human annotations in \autoref{app:extended-sentence-transformers}.

We note that the embedding models used in metrics \metricB\ and \metricD\ are identical to those used in \metricA\ (specifically, the fine-tuned \texttt{BAAI/bge-m3} model), to ensure consistency across semantic evaluations.

\begin{table}[h]
    \centering
    \resizebox{\linewidth}{!}{
    \begin{tabular}{lrr}
        \toprule
        Model & Pearson & Spearman \\
        \midrule
        \multicolumn{3}{c}{\textbf{Cross-lingual}}\\
        \metricA\ Linearized cosine similarity & \textbf{0.66} & \textbf{0.40}\\
        \metricB\ kNN ratio-margin scoring & 0.21 & 0.23 \\ 
        \multicolumn{3}{c}{\textbf{Monolingual}}\\
        \metricC\ BERTScore after translation & 0.22 & 0.16 \\ 
        \metricD\ Sanskrit cosine similarity & 0.42 & 0.25 \\ 
        \bottomrule
    \end{tabular}
    }
    \caption{Pearson and Spearman correlations of different semantic similarity metrics and Human Annotation. {\em Note:} \metricA\, \metricB\ and \metricD\ rely on the same encoder (fine-tuned \texttt{BAAI/bge-m3}).
}% All correlation experiments gave p-values of less than 1E-13.}
    \label{tab:human_corr}
\end{table}
Thus, we choose \metricA\ (the linearized cosine similarity of fine-tuned \texttt{BAAI/bge-m3}) to be the representative metric for semantic similarity.

\subsection{Models considered}
We consider a diverse set of models spanning three broad families to ensure comprehensive coverage and representative evaluation:
(a) \textbf{Raw Translation Models}: These are encoder-decoder models specialized for translation tasks and typically require fine-tuning. They do not support instruction prompts and are not instruction-following. Models in this category include \texttt{IndicBART} ~\cite{indicbart} and \texttt{NLLB200-distilled-1.3B} ~\cite{nllbteam2022languageleftbehindscaling}.
(b) \textbf{Instruction-Following Models}: These are decoder-only models that have been instruction-tuned and can be prompted with task instructions. We use 4-bit quantized versions for efficiency, and they are also suitable for downstream instruction fine-tuning. This category includes \texttt{LLama-3.1-8B}, \texttt{LLama3.2-3B} ~\cite{grattafiori2024llama3herdmodels}, \texttt{Phi-4-14B} \cite{abdin2024phi4technicalreport}, \texttt{Mistral-Nemo-2407-12B} ~\cite{mistral-nemo}, \texttt{Mistral-7B} \cite{jiang2023mistral7b}, \texttt{Qwen2.5-7B} ~\cite{qwen2, qwen2.5}, as well as Indic instruction-tuned models such as \texttt{Airavata-7B} ~\cite{gala2024airavata} and \texttt{Navarasa-9B}.
(c) \textbf{Closed-source Models}: We include a commercial model, \texttt{GPT-4o} ~\cite{mishra2023promptingpseudocodeinstructions}, which is accessible only via few-shot prompting and cannot be fine-tuned.

The prompt used for instruction-finetuned models is provided in Appendix ~\ref{IFT-prompt-appendix}. All training and LoRA hyperparameters are detailed in Appendices ~\ref{Training-hyperparameters} and ~\ref{LoRA-hyperparameters}.

\subsection{Results}

\autoref{tab:syntac-eval} shows that, despite the high zero-shot semantic similarity achieved by the \texttt{NLLB-200-dist-1.3B} model, its syntactic performance on Sanskrit \texttt{Anu\d{s}\d{t}ubh} verse generation remains poor. Notably, applying constrained decoding to the base model improves syntactic well-formedness to 92.61\%, albeit with a marginal degradation in semantic fidelity. When our proposed custom-constrained decoding strategy is applied to the \texttt{NLLB} model fine-tuned on \ourdataset, syntactic correctness reaches 99.86\%, marking a substantial improvement over both the base and simply constrained variants, as well as \texttt{indicBART}.

Interestingly, the unmodified base model without any fine-tuning or decoding constraints achieves the highest semantic similarity among all configurations. This can likely be attributed to the model’s pretraining objective, which centers on prose-to-prose translation, thereby optimizing for semantic fidelity at the expense of structural constraints inherent to metrical verse.

Our experiments reveal a consistent trade-off between syntactic structure and semantic accuracy: while constrained decoding (with or without fine-tuning) enhances syntactic adherence, it does so by compromising semantic alignment. These findings highlight the fundamental traction between form and meaning in neural generation of metrically bound poetic forms.

\begin{table}[ht]
\centering

\resizebox{0.95\linewidth}{!}{
\begin{tabular}{lccrrr}
\toprule

    \textbf{Model} & CD & FT & \textbf{Full \%} & \textbf{Partial \%} & \textbf{Sim} \\
    \midrule
    \multicolumn{6}{c}{\textbf{Raw Translation Models}}\\
    \texttt{IndicBART} & \cross & \tick & 0.0 & 0.63 & 52.60 \\
    \texttt{IndicBART} & \tick & \tick & 73.75 & 74.95 & 54.53 \\
    \texttt{NLLB-dist-1.3B} & \cross & \cross & 0.00 & 0.92 & \textbf{79.33} \\
    \texttt{NLLB-dist-1.3B} & \tick & \cross & \underline{92.61} & 93.53 & \underline{64.96} \\
    \texttt{NLLB-dist-1.3B} & \cross & \tick & 1.69 & 20.76 & 68.33 \\
    \texttt{NLLB-dist-1.3B} & \tick & \tick & \textbf{99.86} & 99.86 & 64.91 \\
    \midrule 
    \multicolumn{6}{c}{\textbf{Instruction Following Models}}\\
    \texttt{Llama3.1-8B} & \cross & \cross & 0.00 & 0.42 & 61.88 \\
    \texttt{Llama3.1-8B} & \cross & \tick & 44.48 & 82.13 & 66.45 \\
    \texttt{Llama3.1-8B} & \tick & \tick & 40.62 & 40.62 & 61.2 \\
    \texttt{LLama3.2-3B} & \cross & \tick & 34.69 & 69.31 & 65.1 \\
    \texttt{LLama3.2-3B} & \tick & \tick & 43.75 & 46.88 & 61.2 \\
    \texttt{Phi-4-14B} & \cross & \tick & \textbf{57.42} & 75.01 & 67.29 \\
    \texttt{Mistral-Nemo-2407-12B} & \cross & \tick & \underline{52.5} & 80.79 & \textbf{68.05} \\ 
    \texttt{Mistral-7B} & \cross & \tick & 43.39 & 66.40 & \underline{67.53} \\
    \texttt{Qwen2.5-7B} & \cross & \tick & 34.41 & 75.58 & 64.97 \\
    \texttt{Airavata-7B} & \cross & \tick & 0.0 & 0.0 & 49.78 \\
    \texttt{Navarasa-9B} & \cross & \tick & 2.08 & 10.55 & 63.08 \\
    
    \midrule
    \multicolumn{6}{c}{\textbf{Closed-source models}}\\
    \texttt{GPT-4o} 1-shot & \cross & \cross & 31.24 &  52.84 &  56.39 \\
\bottomrule
\end{tabular}
}
\caption{Evaluation of different models on the \ourdataset Test dataset. CD : Constrained Decoding; FT : Fine-Tuning on \ourdataset (Train subset). FT in Instruction Following models represents task-specific Instruction Fine-Tuning (IFT).}
\label{tab:syntac-eval}
\end{table}

\begin{table}
\centering
\resizebox{\linewidth}{!}{
\begin{tabular}{lccrrr}
\toprule
    \textbf{Model} & CD & FT & \textbf{Full \%} & \textbf{Partial \%} & \textbf{Sim} \\
    \midrule
    \texttt{GPT-4o (1-shot)} & \cross & \cross & 21.84 & 37.16 & 56.57 \\
    \texttt{NLLB-dist-1.3B} & \tick & \tick & \textbf{99.62} & 99.62 & 61.97 \\
    \texttt{Mistral-Nemo-2407-12B} & \cross & \tick & 52.89 & 77.50 & \underline{66.08} \\
    \texttt{Phi-4-14B} & \cross & \tick & \underline{64.81} & 84.23 & \textbf{66.55} \\
\bottomrule
\end{tabular}
}
\caption{Out of Dataset Sample Evaluation on various models. CD : Constrained Decoding; FT : Fine-Tuning on \ourdataset (Train subset). FT in Instruction Following models represents task-specific Instruction Fine-Tuning (IFT).}

\label{tab:ood}
\end{table}

Instruction-tuned models perform poorly in zero-shot settings but show strong improvements when fine-tuned with Sanskrit-specific prompts. \texttt{Llama-3.1-8B-Instruct} demonstrates this trend, while \texttt{Phi-4-14B} \citep{abdin2024phi4technicalreport} achieves the best performance, with 64.81\% syntactic accuracy and 84.25\% of outputs conforming to partial \texttt{Anu\d{s}\d{t}ubh} structure, alongside a 66.55 semantic similarity score (\autoref{tab:syntac-eval}). We also experimented with Indic-LLMs namely \texttt{Airavata-7B}, \texttt{Navarasa-9B} (LLMs finetuned on Indic data) which did not perform well on the task. Interestingly, our analysis reveals a trade-off between syntactic well-formedness and semantic similarity in Sanskrit \texttt{Anu\d{s}\d{t}ubh} poetry generation, observable across both IFT and CD approaches.

\noindent \textbf{Effect of Constrained Decoding with Instruction-Finetuned Models}
The constrained decoding approach, when applied to the LLaMa model (one of the instruct-finetuned model), produced suboptimal results, in part because the model exhibited a propensity to write English text prior to composing Sanskrit poetry. This issue was absent in the \texttt{NLLB} paradigm, which proficiently employs tags for regulated language creation. As a result, efforts to implement custom decoding directly on \texttt{LLaMa's} outputs failed to produce poetry in \texttt{Anu\d{s}\d{t}ubh}, despite attempts to limit the production to Devanagari tokens.

% Requires: \usepackage{booktabs}
\begin{table*}[ht]
    \centering
    \resizebox{\textwidth}{!}{
    \begin{tabular}{lcccc}
        \toprule
        Model &
        \parbox[c]{2.6cm}{\centering Full Anu\d{s}\d{t}ubh} &
        \parbox[c]{3.7cm}{\centering 32-syllable with \\ Metrical Violations \\ (Partial-Full)} &
        \parbox[c]{3.6cm}{\centering Off-length \\ (30/31/33/34 syllables)} &
        \parbox[c]{2.8cm}{\centering Other Syllable \\ Count Errors \\ ($<$30 or $>$34)} \\
        \midrule
        GPT-4o & 31.24\% & 21.6\% & 27.45\% & 19.7\% \\
        Phi-3-14B (Fine-tuned) & 57.42\% & 17.59\% & 23.93\% & 1.06\% \\
        NLLB-OD (Ours) & 99.86\% & 0\% & 0.14\% & 0\% \\
        \bottomrule
    \end{tabular}
    }
    \caption{Error analysis on \texttt{Anu\d{s}\d{t}ubh} generation, illustrating differences in metrical validity and types of verse-length violations across models.}
    \label{tab:metrical_violation}
\end{table*}

\noindent \textbf{Analysis of GPT-4o's Performance on Structured Poetry Generation} Our analysis reveals three critical failure modes in commercial models, directly addressing the reviewer's concern regarding their metrical limitations. First, syllable-weight constraint dissociation is evident: while 21.6\% of the generated outputs achieved the correct 32-syllable length, they consistently violated the underlying Anu\d{s}\d{t}ubh metrical patterns. This demonstrates a failure by large commercial models to internalize the mandatory coupling between syllable quantity and prosodic weight, a fundamental requirement for generating valid Sanskrit poetry. Second, we observed systematic near-miss errors: 27.45\% of the outputs deviated from the target length by only 1-2 syllables. Specifically, 328 out of the 390 off-length cases were precisely clustered at 31 or 33 syllables. This clustering indicates that the models exhibit incomplete learning of basic structural constraints rather than generating random errors.

\noindent \textbf{Out of Domain Evaluation.} \autoref{tab:ood} shows the inference results of our framework on the OOD dataset, which shows the generalization capability of the models. 

Our proposed constrained decoding framework with \texttt{NLLB} outperforms \texttt{GPT-4o} in terms of generating syntactically correct \texttt{Anu\d{s}\d{t}ubh} verse by a large margin. Instruction fine-tuned models also performed well, with \texttt{Mistral-Nemo} giving the best semantic similarity. We use these two models to perform human evaluation.

\subsection{Human Evaluation}
\begin{table}[h]
    \centering
    \resizebox{\linewidth}{!}{
    \begin{tabular}{lrrrrr}
        \toprule
        Model & CD & FT & \makecell{Syntactic\\Coherence\\(0-1)} & \makecell{Semantic\\Coherence\\(1-5)} & \makecell{Poeticness\\(1-5)} \\
        \midrule
        \texttt{NLLB-dist-1.3B} & \tick & \tick & 0.65 & 2.125 & 2.225 \\
        \texttt{Mistral-Nemo-2407-12B} & \cross & \tick & 0.725 & 3.775 & 3.85 \\
        \bottomrule
    \end{tabular}
    }
    \caption{Human evaluation metrics (avg). The score range for Syntactic Coherence is 0-1 whereas for Semantic Coherence and Poeticness, it is 1-5.}
    \label{tab:human-evaluation}
\end{table}
Human evaluation is crucial in this task, as automated metrics alone cannot fully capture the metrical accuracy or poetic expressiveness of Sanskrit verse. To evaluate the quality of generated poetry, we conducted a human evaluation with three expert annotators, all PhD scholars in Sanskrit. Each annotator assessed non-overlapping subset from the sample of 40 random samples from the test set, rating outputs from two models based on syntactic coherence (adherence to \texttt{Anu\d{s}\d{t}ubh} meter as well as grammaticality), semantic coherence (faithfulness to the input sentence), and poeticness (aesthetic and literary qualities). We refer readers to the \autoref{sec:human_eval} for the detailed design, scoring rubric, and instructions provided to annotators. 

The instruction-tuned \texttt{Mistral-Nemo} model consistently outperforms the constrained \texttt{NLLB} variant in all three aspects. Despite \texttt{NLLB}'s, technically high structural compliance with the \texttt{Anu\d{s}\d{t}ubh} meter, it lags behind slightly in syntactic coherence due to grammatical incoherence, as judged by human evaluators. This gap highlights a key limitation of hard-constrained decoding, as it enforces formal metrical correctness but often at the expense of fluency and naturalness.

\texttt{Mistral-Nemo} surpasses \texttt{NLLB} in semantic coherence, which is consistent with automated evaluation metrics. 

\texttt{Mistral-Nemo}'s superior performance in poeticness can be explained by the model’s strong transfer learning capabilities, augmented by fine-tuning on literary-style instruction data and curated prompts that promote stylistic and creative fluency.

\iffalse

\paragraph{Criteria} We utilize the evaluation framework established by \cite{zhang-lapata-2014-chinese}, wherein human evaluators assign ratings to poems on a scale of 1 to 10 across four principal dimensions:

\begin{itemize}
    \item \textit{Fluency}: exhibit grammatical and syntactically correct structure. 
    \item \textit{Coherence}: maintain thematic organization.
    \item \textit{Meaningfulness}: communicate a significant message. 
    \item \textit{Poeticness}: exhibit the characteristics of poetry.
\end{itemize}
\fi

\subsection{Ablation Analysis}
We perform an ablation analysis focusing on various hyper-parameters such as the choice of $k$ as well as sampling strategy in the constrained decoding algorithm.

\noindent \textbf{On the role of $k$.} $k$ represents the number of tokens with the highest probabilities under consideration - in order to check whether they satisfy metrical constraints. We set this to $k= 25$, but also test what happens when we have \textit{too few} or \textit{too many} token options. From \autoref{tab:ablation-k} in \autoref{app:ablations}, we can see that $k=25$ still performs the best overall.% These are further inferred .

\noindent\textbf{Different Sampling Strategies.} We experiment with different sampling strategies after applying constrained decoding. \autoref{tab:sampling} of \autoref{app:ablations} summarizes the results for different sampling strategies such as greedy, multinomial, nucleus, and top-k. We find that greedy decoding performs the best, both structurally and semantically. 
\label{p:ablation-sampling}

\noindent \textbf{Effect of Incorporating Constrained Decoding Algorithm in the Prompt.}
We perform ablation analysis of incorporating constrained decoding algorithm in the prompt for generating Sanskrit \texttt{Anu\d{s}\d{t}ubh} Poetry. We experimented by including \autoref{alg:constrained_decoding} in the prompt provided to \texttt{GPT-4o} (see \autoref{image/IFT-prompt}). \autoref{tab:prompt+cd} shows that it does not improve performance further and in fact led to a decrease, reaffirming the necessity of our decoding-time constraints.

\noindent\textbf{Latency and Throughput Analysis of CD.} We analyze the effect our custom decoding strategy (Constrained Decoding) has on the speed of generation. As expected, there is a decrease caused by \textit{non-parallelized decoding} and the overhead of the \texttt{LogitsProcessor} interface with the \texttt{transformers} library. 
Notably, 90\% of the time is spent inside the \texttt{skrutable} library~\cite{skrutable}, syllabifying the received generated tokens and available options at each step, signalling that the actual algorithm and regex checking/length checking don't add much. To somewhat alleviate this, we do add a Least-Recently-Used (LRU) cache of size 1000 to the syllable scanning, which helps, as shown in \autoref{tab:latency}.

\begin{table}[h]
    \centering
    \resizebox{\linewidth}{!}{
    \begin{tabular}{lrrr}
        \toprule
        Model & CD & Latency (s) & Throughput (tok/s) \\
        \midrule
        \texttt{NLLB-dist-1.3B} & \cross & 0.041 & 932.8 \\
        \texttt{NLLB-dist-1.3B} & \tick & 0.280 & 147.6 \\
        \texttt{NLLB-dist-1.3B + cache} & \tick & 0.219 & 181.4 \\
        \midrule
        \texttt{Llama3.1-8B-Instruct} & \cross & 0.936 & 73.7 \\
        \texttt{Llama3.1-8B-Instruct} + cache & \tick & 2.550 & 30.0 \\
        \bottomrule
    \end{tabular}
    }
    \caption{Impact of CD on the Latency and Throughput of generation.}
    \label{tab:latency}
\end{table}

\section{Generalization of Constrained Decoding for Poetry Generation}
The performance of the constrained decoding approach is tabulted in \autoref{tab:other-chanda-lang}. The subsections dive deeper into generalization across Sanskrit meter and other languages.
\begin{table}
    \centering
    \resizebox{0.482\textwidth}{!}{
    \begin{tabular}{lcccc}
    \toprule
 & \textbf{Sanskrit}  & \textbf{Sanskrit} & \textbf{Awadhi} & \textbf{Bengali}\\
\textbf{Evaluation} & \textbf{(Anu\d{s}\d{t}ubh)} & \textbf{(Tri\d{s}\d{t}ubh)} & \textbf{(Doha)} & \textbf{(Payar)}\\
& \textbf{[1421]} & \textbf{[30]} & \textbf{[30]}& \textbf{[30]}\\
    \midrule
        Full Compliance & 99.86 & 96.67 & 86.67& 83.33\\
        Partial Compliance & 99.86 & 96.67 & 90.0& 83.33\%\\
        Semantic Similarity &  64.91& 68.39 & \cross & \cross \\
         
         \bottomrule
    \end{tabular}
    }
    \caption{Performance of constrained decoding approach for different languages and it's respective chanda(s). [.] denotes the sample size, (.) contains the chanda name. }
    \label{tab:other-chanda-lang}
\end{table}
\subsection{Generalization Across Sanskrit Meters} 
The constrained decoding approach exhibits robust extensibility across diverse Sanskrit meters, facilitated by a modular design encompassing regex compilation and length constraints (\autoref{alg:constrained_decoding}, Steps 5 \& 7). To substantiate this claim, we evaluated the fine-tuned NLLB model on Tri\d{s}\d{t}ubh Chandas (a meter comprising four quarters of eleven syllables each). Using a test set of 30 English translations from the Bhagavadgita, the model demonstrated high structural fidelity, achieving 96.67\% full compliance and 96.67\% partial compliance. The semantic similarity was measured at 68.39. These results confirm the effective generalization of our methodology across varied metrical frameworks inherent to Sanskrit.

\subsection{Generalization to Other Languages} 
The proposed constrained decoding approach generalizes beyond Sanskrit to other Indic languages exhibiting analogous prosodic organization. To assess this, we applied the method to Awadhi and Bengali poetic forms. For Awadhi\footnote{\href{https://en.wikipedia.org/wiki/Awadhi_language}{https:\/\/en.wikipedia.org/wiki/Awadhi\_language}}, we targeted Doha Chandas—defined by 13 m\={a}tr\={a}s in the first and third quarters and 11 m\={a}tr\={a}s in the second and fourth, totaling 48 m\={a}tr\={a}s. Using the base NLLB-600M model on 30 Doha samples drawn from R\={a}macaritam\={a}nasa, the system achieved 86.67\% full compliance and 90.0\% partial compliance. Bengali experiments used P\={a}yar—comprising two lines of 14 syllables each—with 30 samples sourced from \'Sr\={i} Caitanya-carit\={a}m\d{r}ta. Under identical modeling conditions, NLLB-600M attained 83.33\% full and partial compliance. These findings underscore that the core notion of laghu–guru weighting (1 vs 2 m\={a}tr\={a}s) extends naturally across Indo-Aryan and Dravidian meters, facilitating rapid adaptation to additional regional traditions, including Telugu, Tamil, and Kannada.

\section{Conclusion}
We introduced \ourdataset, a benchmark for generating Sanskrit poetry from English in the \texttt{Anu\d{s}\d{t}ubh} meter, and evaluated a range of open and closed models. Constrained decoding and instruction fine-tuning emerged as effective strategies: \texttt{NLLB-dist-1.3B} attains 99.86\% metrical accuracy with constrained decoding, while \texttt{Phi-4-14B} achieves 57.42\% fully valid verses after instruction tuning. Human judgments confirm that instruction-tuned models better capture poetic form and style, and out-of-domain testing demonstrates that constrained decoding outperforms larger commercial systems such as \texttt{GPT-4o} in enforcing prosodic rules.

Preliminary experiments across other chandas and languages suggest that constrained decoding generalizes reliably beyond Sanskrit and \texttt{Anu\d{s}\d{t}ubh}. Overall, our work provides initial baselines for English-to-Sanskrit poetry generation with metrical control and points toward expanding datasets, broader meter coverage, improved semantic metrics, and scaling tuning strategies to multilingual and larger models.

\section{Limitations}
We could not evaluate larger LLMs due to limited availability of compute resources (single GPU); hence, we selected a few representative smaller models for our experiments. Furthermore, while our methods remain meter-agnostic, we have not tested them with other meters. Also, while the proposed method is general, we have not explored generating Sanskrit poetry from inputs in other languages than English. 

One limitation of CD is it's application alongside IFT models. This is problematic because the IFT models might generate some generic English/part of the instruction before actually generating the verse. Constrained Decoding skews the output distribution (and increases perplexity) by not allowing this to happen and hence does not perform well. This problem is exacerbated by the BPE (byte pair encoding) of the sentencepiece tokenizer, which breaks some Sanskrit tokens into a pair of two tokens ids. But, CD is limited to only predicting one token from the \texttt{LogitsProcessor} interface and hence cannot verify/predict these tokens.

As a preliminary probe of cross-chandas and cross-lingual generalization, we evaluated constrained decoding on a small sample of 30 instances drawn from one representative meter each in Awadhi and Bengali. Given this limited scope, our findings should be interpreted as indicative rather than comprehensive. Future work will expand these experiments with larger datasets and extended meter coverage, and will additionally assess the effects of instruction fine-tuning in these settings.

\section{Acknowledgement}
We gratefully acknowledge Sinchana Bhat and Archana Bhat, Classical Sanskrit Linguists at C-DAC Bangalore, for their invaluable contributions to annotation and human evaluation. We also thank Mohan Kumar, Junior Research Associate at the Indian Institute of Technology (BHU) Varanasi, and Prabhat Bhardwaj, PhD Scholar at the University of Delhi, for their support in human evaluation.

This work was supported in part by the National Language Translation Mission (NLTM): Bhashini project of the Government of India.

\bibliography{arxiv-manuscript}

\appendix
\onecolumn

\appendix

\begin{center}
    \Large \textbf{Appendix}
\end{center}

\section{Extended background on Chandas}
\label{app:ext-chandas}
Meters in Sanskrit are broadly classified into:
\begin{enumerate}
    \item \texttt{Var\d{n}av\d{r}tta}: Based on a fixed number and sequence of syllables (laghu and guru).
    \item \texttt{M\={a}tr\={a}vṛtta}: Based on the count of \texttt{m\={a}tr\={a}s} (morae or units of time), regardless of the syllable count.
    \item \texttt{Samav\d{r}tta, Ardhasamav\d{r}tta, Vi\d{s}amav\d{r}tta}: Subcategories based on uniformity or variation across the \texttt{p\={a}das}.
\end{enumerate}
The work by \newcite{melnad2013meter} explains the meters in detail. \texttt{Anu\d{s}\d{t}ubh} is an example of \texttt{Samav\s{r}tta} meter. 

The patterns for the meters is specified through the use of \texttt{Ga\s{n}as}. \texttt{Ga\s{n}as} denote patterns of three syllables, combining light and heavy syllables. There are eight types of \texttt{ga\s{n}as}, represented mnemonically as: 
\begin{multicols}{2}
\begin{itemize}
    \item ya: laghu-guru-guru
    \item ra: guru-laghu-guru
    \item ta: guru-guru-laghu
    \item bha: guru-laghu-laghu
    \item ja: laghu-guru-laghu
    \item sa: laghu-laghu-guru
    \item ma: guru-guru-guru
    \item na: laghu-laghu-laghu
\end{itemize}
\end{multicols}

\section{Additonal Details}
\label{p:details}
\subsection{Sanskrit Poetry Generation Prompt}
\label{IFT-prompt-appendix}
\begin{figure}[h!]
\centering
\resizebox{!}{10cm}{
    \includegraphics{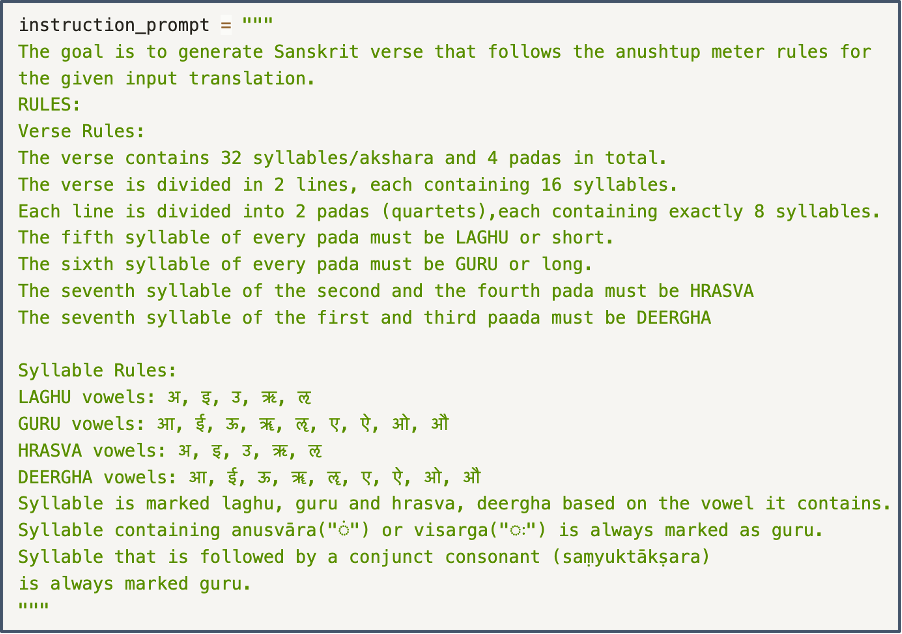}
    }
    \caption{Prompt used for Instruction fine-tuning, providing necessary grammatical and \texttt{Anu\d{s}\d{t}ubh} meter rules}
    \centering
    \label{image/IFT-prompt}
\end{figure}
Please refer to \autoref{image/IFT-prompt} for the prompt used for zero-shot / few-shot and instruction fine-tuning.

\subsection{Definition of Anu\d{s}\d{t}ubh in original Sanskrit}
\label{anushtubh-def}
This metrical pattern is traditionally encoded in the following Sanskrit mnemonic verse:\\
{{\skt (\ZM{lBl0FI0PE};+ea;k\ZH{-12}{e} :Sa;S2+m,a gua:r8 :]ea;yMa .sa;vRa:t3a
l+.Gua :pa:\ZM{BNz0dcRfA}*.a;ma;m,a \ZS{12}@A \\
;Y2a;d\ZM{oau};.ca;tua;Spa;a;d;ya;eaH\ZS{4} :h\ZM{ojr};s1vMa .sa;p1a;mMa
d\ZH{-6}{i6};a;GRa;ma;nya;ya;eaH\ZS{4} \ZS{12}@A\ZS{6}@A}}

\footnote{Another definition of Anu\d{s}\d{t}ubh: \\ {{\skt :pa:\ZM{BNz0dcRfA}*.a;mMa l+.Gua .sa;vRa:t3a .sa;p1a;mMa
;Y2a;d\ZM{oau};.ca;tua;TRa;ya;eaH\ZS{4} \ZS{12}@A \\
gua:r8 :Sa;S2\ZH{-6}{M} ..ca .sa;veRa;Sa;a;m,a O;;ta;.c%
\ZM{0PC0dlLAI}*:+ea;k+:s1ya l+.[a;Na;m,a \ZS{12}@A\ZS{6}@A}}
}\\

\textbf{Translation:}\\
``The fifth syllable is always light;\\
The seventh is light in the second and fourth lines;\\
The sixth is always heavy $—$ this defines the characteristics of the \texttt{\'{s}loka}.''

\subsection{Human Annotation Scoring for Metric Evaluation}
\label{app:human-annotation-metriceval}
Refer to \autoref{sec:semantic-similarity-methods}

% \vspace{0.5\baselineskip}
\noindent\fbox{
\begin{minipage}{\dimexpr\linewidth-2\fboxsep-2\fboxrule\relax}    
Score Metric: \\
5: Meaning fully preserved with nuanced fidelity\\
4: Mostly preserved; minor omissions or shifts\\
3: General sense preserved, but key elements missing\\
2: Only a vague resemblance to input meaning\\
1: Completely unrelated
\end{minipage}
}
% \vspace{0.5\baselineskip}

\subsection{Constrained Decoding Algorithm}
\begin{algorithm*}[h!]
\caption{Constrained Decoding for Sanskrit Poetry Generation
\\ %\pg{In the algo, the sampling part is not mentioned.} {\color{cyan} - Fixed}
}\label{alg:constrained_decoding-app}
\SetKwInOut{Input}{Input}
\SetKwInOut{Output}{Output}

\Input{Pre-trained LLM, input context, metrical constraints (e.g.,  rules) in the form of pre-compiled regular expression filters}
\Output{Generated Sanskrit poetry adhering to metrical constraints}

$|V| \gets$ Vocab size of the LLM \quad
$k \gets 25$

\While{not end of generation}{
    
    \textbf{Step 1:} Generate token probabilities from the LLM\quad
    $P \in \mathbb{R}^{|V|} \gets \textbf{LLM}(\text{input})$\;
    \textbf{Step 2:} Sample top-$k$ possible tokens\quad
    Top-$k$ Indices: $I_k \gets \texttt{topk}(P, k)$\;
    \For{each token at index $I_i$ in top-$k$ token indices $I_k$}{
        Token $T_i \gets \text{vocab}[I_i], \quad$
        $O_{\text{temp}} \gets \{\text{...input tokens}\}$\;
        \textbf{Step 3:} Temporarily add $T_i$ to output $O_{\text{temp}}$
        $\quad O_{\text{temp}} \gets O_{\text{temp}} || T_i$\;
        \textbf{Step 4:} Calculate the new syllable weighting for $O_{\text{temp}}$
        $\quad  w_{\text{temp}} \gets \text{syllable-weights}(O_{\text{temp}})$\;
        \textbf{Step 5:} Match $w_{\text{temp}}$ against pre-compiled regex filters of the same length\;
        \If{$w_{\text{temp}}$ does not match the filters}{
            \textbf{Step 6:} $P[I_i] \gets -\inf$
        }
        \textbf{Step 7:} Check length rules (e.g., end of pada or verse)\;
        % \If{token satisfies length rules}{
        %     \textbf{Step 8:} Break the loop and return $T_i$ as the chosen token.\;
        % }
        % \Else{
        % }
    }
    \If{no token matches the constraints}{
        \textbf{Step 8:} Increase $k$ and repeat the process from Step 1\;
    }
    \textbf{Step 9:} Return modified probabilities for further sampling (greedy/nucleus/etc.)
}

\end{algorithm*}
Refer to \autoref{alg:constrained_decoding-app}

\section{Training Hyperparameters for Instruction-Finetuned Models }
\label{Training-hyperparameters}

\begin{itemize}[leftmargin=1.5cm]
    \item \textbf{Maximum Sequence Length:} 1024
    \item \textbf{Warmup Steps:} 5
    \item \textbf{Learning Rate:} $2 \times 10^{-4}$
    \item \textbf{Optimizer:} \texttt{adamw\_8bit}
    \item \textbf{Weight Decay:} 0.01
    \item \textbf{Learning Rate Scheduler Type:} \texttt{linear}
\end{itemize}

\section{LoRA Hyperparameters for Instruction-Tuned Models}
\label{LoRA-hyperparameters}

\begin{itemize}[leftmargin=1.5cm]
    \item $r = 16$
    \item \textbf{Target Modules:} \texttt{["q\_proj", "k\_proj", "v\_proj", "o\_proj", "gate\_proj", "up\_proj", "down\_proj"]}
    \item \textbf{LoRA Alpha:} 16
    \item \textbf{LoRA Dropout:} 0
    \item \textbf{4-Bit Quantization:} \emph{If used}
\end{itemize}

\section{Extended evaluation of \metricA\ on Sentence Transformer Models}
\label{app:extended-sentence-transformers}
Refer to \autoref{tab:appendix-human_corr}.
\begin{table}[h]
    \centering
    % \resizebox{\linewidth}{!}{
    \begin{tabular}{lrr}
        \toprule
        Model & Pearson & Spearman \\
        \midrule
        paraphrase-multilingual-MiniLM-L12-v2 & 0.63 & 0.38\\
        LaBSE & 0.64 & 0.38\\
        Alibaba-NLP/gte-multilingual-base & 0.65 & 0.38\\
        BAAI/bge-m3 & \textbf{0.66} & \textbf{0.40}\\
        FacebookAI/xlm-roberta-large & 0.64 & 0.38\\
        % Facebook/MEXMA& 560& {\color{red} TODO} \\
        %Avg & 0.67 & 0.40\\
        \bottomrule
    \end{tabular}
    % }
    \caption{Pearson and Spearman correlations of \textbf{Sentence Transformer Models} for metric \metricA (linearized cosine similarity) (\autoref{sec:semantic-similarity-methods}) and Human Evaluation.}
    \label{tab:appendix-human_corr}
\end{table}

\section{Ablation: Constrained Decoding}
\label{app:ablations}
\subsection{Effect of pseudo-code in prompt}

\begin{table}[h]
\centering

\begin{tabular}{lrr}
\toprule
    \textbf{Model} & \textbf{Full \%} & \textbf{Partial \%}  \\
    \midrule
    GPT-4o (1-shot) & \textbf{31.24} & \textbf{52.84} \\
    GPT-4o (1-shot + CD) & 26.46 & 48.35 \\

\bottomrule
\end{tabular}
\caption{Evaluation of GPT-4o on the  Test dataset. CD : Constrained Decoding Algorithm as pseudo-code in prompt.}

\label{tab:prompt+cd}
\end{table}

\subsection{Effects of \texorpdfstring{$k$}{k}}

\begin{table}[h]
    \centering
    \begin{tabular}{lrrrrr}
    \toprule
       
       $k$ & \textbf{Full \%} & \textbf{Partial \%} & \textbf{Sim} & \textbf{Latency (s)} & \makecell{\textbf{Throughput}\\ (tok/s)} \\
    \midrule
       10 & 99.30 & 99.93 & 64.91 & 0.160 & 248.3 \\
       25 & 99.86 & 99.86 & 64.91 & 0.219 & 181.4 \\
       50 & 99.93 & 99.93 & 64.92 & 0.413 & 96.1 \\
       100 & 99.93 & 99.93 & 64.92 & 0.741 & 53.604 \\
    \bottomrule
    \end{tabular}
    \caption{Effect of increasing and decreasing $k$ on constrained decoding. The model used was NLLB-dist-1.3B-FT. The metrics mean the same as in \autoref{tab:syntac-eval}}
    \label{tab:ablation-k}
\end{table}

We test what happens when we have \textit{too few} or \textit{too many} token options in constrained decoding. From \autoref{tab:ablation-k}, we can see that k = 25 still performs the best overall. 

\begin{itemize}
    \item \textbf{Structurally}, a moderate value of $k$ gives the best results. This is because when $k$ is low, there may not be enough options to satisfy metrical criteria.
    \item \textbf{Semantically}, it is intuitively best to choose the highest probability option - which may not be chosen when $k$ is high, because of either constraints not being met, or sampling. The metrics show a clear trend of decreasing similarity with increasing $k$.
\end{itemize}

\subsection{Sampling Strategies}
\begin{table}[h]
    \centering
    % \resizebox{\linewidth}{!}{
    \begin{tabular}{lrrr}
    \toprule
       Sampling strategy & \textbf{Full \%} & \textbf{Partial \%$^2$} & \textbf{Sim$^3$} \\
    \midrule
       Greedy & \textbf{99.86} & 99.86 & \textbf{64.91} \\
       Multinomial (Temperature $T=0.7$) & 72.56 & 97.19 & 64.03 \\
       Nucleus/Top-p ($p=0.9$) & 62.35 & 95.07 & 63.41 \\
       Top-k ($k=10$) & 79.80 & 97.33 & 63.42 \\
       Contrastive Search ($\alpha=0.6,k=4$) & 92.12 & 99.16 & 64.25 \\
    \bottomrule
    \end{tabular}
    % }
    \caption{Effect of changing the sampling strategy on constrained decoding. The model used was NLLB-dist-1.3B-FT. The metrics mean the same as in \autoref{tab:syntac-eval}}
    \label{tab:sampling}
\end{table}

As mentioned in \autoref{p:ablation-sampling}, we explore the effects choosing a different sampling strategy has, after applying constrained decoding. These are shown in \autoref{tab:sampling}.

\section{Dataset Examples}
\label{sec:dataset-examples}
We present some extracts from {\ourdataset} in \autoref{tab:dataset-samples}.
\begin{table}[!thb]
    \centering
    \begin{tabular}{m{7.5cm}|m{7.5cm}}
    \toprule
        \textbf{Input (English Prose)} & \textbf{Output (Sanskrit Poetry)} \\
        \midrule
        Effulgent Rama looked at Khara who stood with a mace in hand minus his chariot and said to him first in a gentle voice and then harshly & {{\skt Ka:=\ZH{-6}{M} tua ;Y2a;va:=+TMa .=+a;ma;ea
ga;d;a;pa;a;Y5a;Na;ma;va;\ZH{0}{i0//////Y7}a;s1Ta;ta;m,a \ZS{12}@A 
m\ZV{2}{x}a;d\ZH{-10}{u};pUa;vR2a ma;h;a;tea:ja;aH\ZS{4} :pa:r8+:SMa
va;a;ky2+.a;ma;b.ra;vi6a;a;t,a \ZS{12}@A\ZS{6}@A}} \\\hline
        Having witnessed the destruction of those regions earned by him through asceticism, the son of Jamadagni left for the best of mountains Mahendra. & {{\skt .sa h;ta;a;n,a d\ZH{-4}{\ZV{2}{x}};Zya .=+a;mea;Na
.s1va;Ma;l2+:ea;k+:Ma;s1ta;pa;sa;a;Y6a:jRa;ta;a;n,a \ZS{12}@A 
.ja;a;ma;d;\ZM{0NgFRI0FnIFe}+;ay1a;ea .ja;ga;a;ma;a;Zua
ma;h\ZH{-6}{e};nd\ZP{-8}{-4}{@R}\ZH{-6}{M} :pa;vRa;ta;ea:t1a;ma;m,a
\ZS{12}@A\ZS{6}@A}} \\ \hline
        Again, and again looking back in fear, covertly, bewildered, crushing one another, they went into the Lanka city. & {{\skt A;nya;ea;nyMa :pra;ma;ma;nTua;s1tea ;Y2a;va;Y2a;va;Zua;nRa;ga:=%
\ZH{-6}{M}+Ba;ya;a;t,a \ZS{12}@A 
:p\ZV{4}{x}a;S2+ta;s1tea tua .sa;mmUa;Q+aH\ZS{4} :prea;[a;ma;a;Na;a;mua;h%
\ZH{-8}{\ZV{-8}{u}};mRua;h\ZH{-8}{\ZV{-8}{u}}H\ZS{4} \ZS{12}@A\ZS{6}@A}} \\
    \bottomrule
    \end{tabular}
    \caption{Samples from \ourdataset dataset}
    \label{tab:dataset-samples}
\end{table}

\section{Examples of Input-Output pairs from Different Models}
\label{sec:outputs}
For the following input, model generated Sanskrit poetry is shown in \autoref{tab:output-examples}.

\textbf{Input:} 
\begin{quote}
Effulgent Rama looked at Khara who stood with a mace in hand minus his chariot and said to him first in a gentle voice and then harshly
\end{quote}

\begin{table}[!thb]
    \centering
    \begin{tabular}{l|m{10cm}}
    \toprule
        \textbf{Model}  & \textbf{Generated Sanskrit poetry} \\
        \midrule
        \texttt{Mistral-Nemo-2407-12B} &  {{\skt tMa ga;d;a;pa;a;Y5a;Na;nMa d\ZH{-4}{\ZV{2}{x}};S1\ZM{pfuffE};a
;Y2a;va;h\ZH{-6}{i6};a;nMa .=+Ta;sa:t1a;ma;m,a \ZS{12}@A 
:pra;ea;va;a;.ca :pra;Ta;mMa .=+a;ma;ea m\ZV{2}{x}a;d\ZH{-6}{M}\ZH{-10}{u}
va;a;ky2+.a\ZH{-6}{M} ta;ta;s1ta;taH\ZS{4} \ZS{12}@A\ZS{6}@A}} \\ \hline
        \texttt{NLLB-dist-1.3B} (CD)  & {{\skt .=+Ta;a;\ZH{0}{i0//////Y7}a;t[a;prMa tua .=+a;ma;ea .=;j2va;l+.n,a
Ka;l\ZH{-6}{u} .sa;mi6a;a;i8a:=+ta;m,a \ZS{12}@A 
o+.va;a;.ca Ba;a;gRa;vMa .=+a;ma;ea ;Dya;a;ya;n,a ;%
\ZH{0}{i0///Y7}a;s1:t4a;ya;Y6a;ma;va;a;.ca;l\ZH{-8}{e} \ZS{12}@A\ZS{6}@A}} \\ \hline
        \texttt{GPT-4o} (1-shot) &  {{\skt .=+Ta;h\ZH{-6}{i6};a;nMa Ka:=\ZH{-6}{M} d\ZH{-4}{\ZV{2}{x}};S1%
\ZM{pfuffE};a .=+a;ma;ea m\ZV{2}{x}a;d\ZM{oau};a .ja;ga;a;d ta;m,a \ZS{12}@A
:pa;(\ZM{rac0FI}a;a;t,a tMa k\ZH{-12}{u}+:Y2a;pa;ta;ea va;a;ky2+.a\ZH{-6}{M}
o+.va;a;.ca .tea:ja;sa;a yua;taH\ZS{4} \ZS{12}@A\ZS{6}@A}} \\
    \bottomrule
    \end{tabular}
    \caption{Sample output from different models for the input, ``Effulgent Rama looked at Khara who stood with a mace in hand minus his chariot and said to him first in a gentle voice and then harshly".}
    \label{tab:output-examples}
\end{table}

\section{Human Evaluation}
\label{sec:human_eval}
We conducted a structured human evaluation study to complement our automated metrics and assess the linguistic and poetic quality of the generated Sanskrit verses. Three expert annotators, each holding a PhD in Sanskrit, were engaged to assess the outputs of two models (\texttt{NLLB-dist-1.3B-FT+CD} and \texttt{Mistral-Nemo-2407-12B} ) across 40 samples. The evaluation dataset was allocated to the annotators in a non-overlapping manner, with sample sizes of 15, 15, and 10, respectively..

Each evaluation sample consisted of:

\begin{itemize}
    \item The input English sentence.
    \item The ground truth Sanskrit verse composed in  Chandas
    \item The verses generated by \texttt{NLLB-dist-1.3B-FT+CD} and \texttt{Mistral-Nemo-2407-12B}.
\end{itemize}

Annotators were provided with detailed instructions outlining the evaluation criteria, along with a few illustrative examples to calibrate their assessments. They were asked to score each model’s output independently on the following three dimensions using a 0-1 for Syntactic Coherence and 1-5 Likert scale (1 being poor and 5 being excellent) for rest of the metrics:

\textbf{Syntactic Coherence}: Whether the generated verse conforms to the structural constraints of  Chandas, including syllabic count and metrical pattern.

\textbf{Semantic Coherence}: The degree to which the generated verse semantically aligns with the input English sentence.

\textbf{Poeticness}: The extent to which the output demonstrates poetic features such as use of rasa (aesthetic flavor), \texttt{alank\={a}ra} (figures of speech), rhyming or alliteration, and elevated or ornate vocabulary.

The annotators were also informed about the intended use of their evaluations in the analysis section of this paper. Inter-annotator agreement is not applicable here due to the disjoint set of samples assigned to each annotator. Nonetheless, given the annotators' domain expertise and the clear evaluation rubric, the scores provide meaningful insights into the relative strengths and weaknesses of the two models in terms of generating metrically accurate, semantically faithful, and aesthetically compelling Sanskrit poetry.
\end{document}